\documentclass{article}

\usepackage{arxiv}
\usepackage{amsmath}
\usepackage[utf8]{inputenc} % allow utf-8 input
\usepackage[T1]{fontenc}    % use 8-bit T1 fonts
\usepackage{hyperref}       % hyperlinks
\usepackage{url}            % simple URL typesetting
\usepackage{booktabs}       % professional-quality tables
\usepackage{amsfonts}       % blackboard math symbols
\usepackage{nicefrac}       % compact symbols for 1/2, etc.
\usepackage{microtype}      % microtypography
\usepackage{cleveref}       % smart cross-referencing
\usepackage{graphicx}
\usepackage{natbib}
\usepackage{multirow}
\usepackage{bbm}

\title{Semi-Supervised Semantic Segmentation with Uncertainty-guided Self Cross Supervision}

\date{}

\author{ 
	Yunyang Zhang\textsuperscript{1}  \hspace{2mm} Zhiqiang Gong\textsuperscript{1}  \hspace{2mm} Xiaohu Zheng\textsuperscript{2} \hspace{2mm} Xiaoyu Zhao\textsuperscript{1} \hspace{2mm} Wen Yao\textsuperscript{1}
	\\
	Chinese Academy of Military Science\textsuperscript{1}\\
	National University of Defense Technology\textsuperscript{2}\\
	zhangyunyang17@csu.ac.cn
}

\renewcommand{\shorttitle}{\textit{arXiv} Template}

\begin{document}
\maketitle

\begin{abstract}
	As a powerful way of realizing semi-supervised segmentation, the cross supervision method learns cross consistency based on independent ensemble models using abundant unlabeled images. However, the wrong pseudo labeling information generated by cross supervision would  confuse the training process and negatively affect the effectiveness of the segmentation model.
	Besides, the training process of ensemble models in such methods also multiplies the cost of computation resources and decreases the training efficiency.
	To solve these problems, we propose a novel cross supervision method, namely uncertainty-guided self cross supervision (USCS). In addition to ensemble models, we first design a multi-input multi-output (MIMO) segmentation model which can generate multiple outputs with shared model and consequently impose consistency over the outputs, saving the cost on parameters and calculations. On the other hand, we employ uncertainty as guided information to encourage the model to focus on the high confident regions of pseudo labels and mitigate the effects of wrong pseudo labeling in self cross supervision, improving the performance of the segmentation model.
	Extensive experiments show that our method achieves state-of-the-art performance while saving 40.5\% and 49.1\% cost on parameters and calculations. The code is available on GitHub.\footnote{The code will be released after the publication.}
\end{abstract}

% keywords can be removed

\section{Introduction}
Semantic segmentation is a significant fundamental task in computer vision and has achieved great advances in recent years. Compared with other vision tasks, the labeling process for semantic segmentation is much more time and labor consuming. Generally, tens of thousands of samples with pixel-wise labels are essential to guarantee good performance for such a known data-hungry task. However, the high dependence of large amounts of labeled data for training would undoubtedly restrict the development of semantic segmentation.
Semi-supervised semantic segmentation, employing limited labeled data as well as abundant unlabeled data for training segmentation models, is regarded as an effective approach to tackle this problem, and has achieved remarkable success for the task \cite{lai2021semi,liu2021perturbed,ouali2020semi,gong2019cnn,gong2020statistical}.

Advanced semi-supervised semantic segmentation methods are mainly based on consistent regularization. It is under the assumption that the prediction for the same object with different perturbations, such as data augmentation for input images \cite{french2019semi,lai2021semi}, noise interference for feature maps \cite{ouali2020semi} and the perturbations from ensemble models \cite{chen2021semi,ke2020guided}, should be consistent.
Among these perturbations, the one through ensemble models usually provide better performance since it can learn the consistent correlation from each other adaptively, which is so-called cross supervision \cite{chen2021semi}.
The earnings of a single model acquired from unlabeled images can be improved by cross supervision between models achieved by forcing consistency of the predictions.

However, there still exist two drawbacks during cross supervision of semi-supervised segmentation. Firstly, the cost of time and memory for cross supervision is usually multiplicatively increased due to the parallel training of ensemble models with different model architectures or different initializations. Secondly, the wrong pseudo labeling of unlabelled samples generated by the prediction of ensemble models would confuse the training process and make false propagation from one model to others.

\begin{figure}
	\centering
	\includegraphics[width=0.8\linewidth]{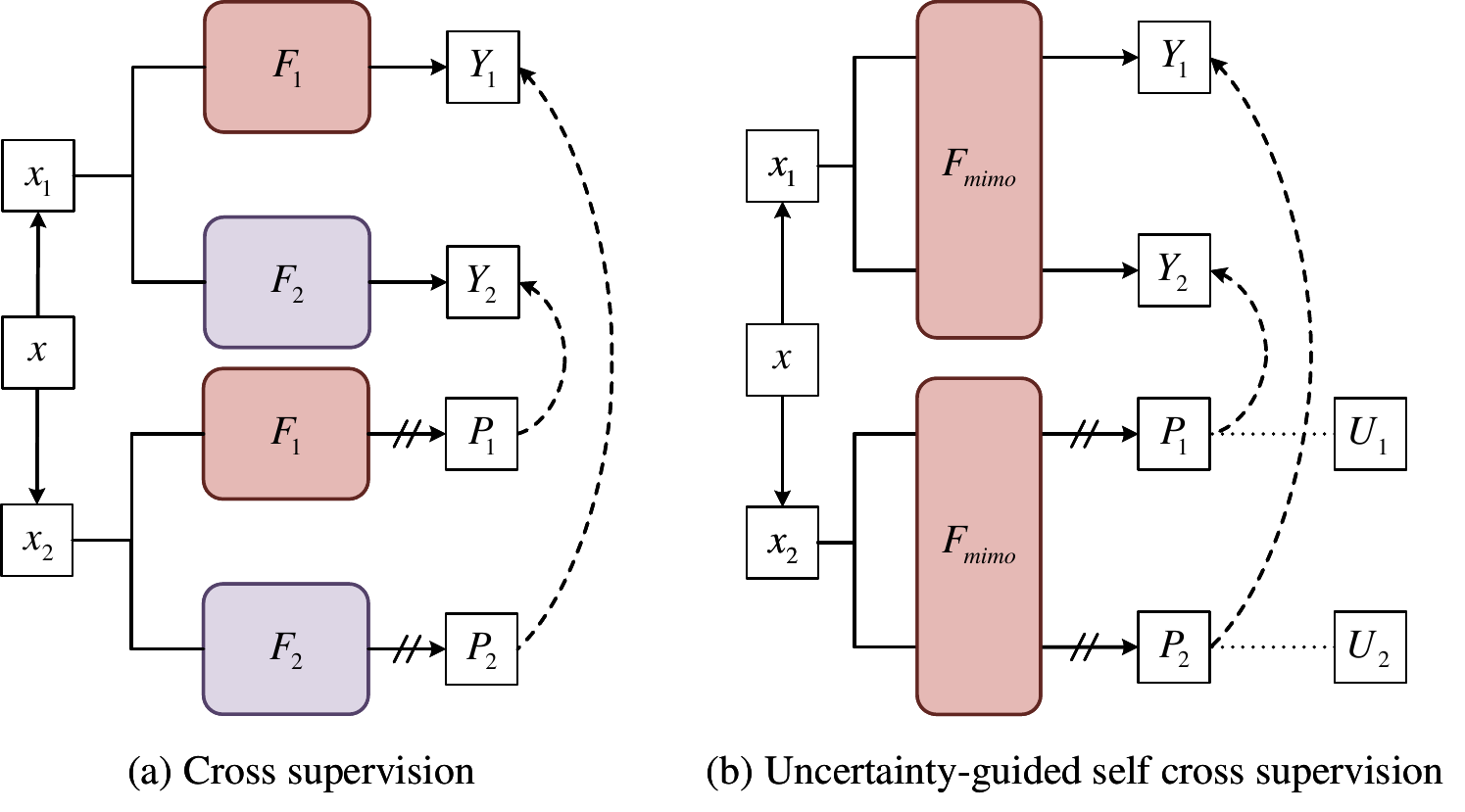}
	\caption{\textbf{Illustrating the architectures} for (a) cross supervision \cite{chen2021semi} and (b) our method uncertainty-guided self cross supervision. In our approach, $Y_1$ and $Y_2$ mean the predictions from a multi-input multi-output model $F_{mimo}$; $P_1$ and $P_2$ mean the pseudo labels for $Y_1$ and $Y_2$; $U_1$ and $U_2$ mean the uncertainty of $P_1$ and $P_2$, respectively.}
	\label{fig:1}
\end{figure}

To address the above two issues, we propose an improved cross supervision method, called uncertainty-guided self cross supervision(USCS). 
The comparisons between cross supervision and our USCS is shown in Fig.\ref{fig:1}.

Specifically, for the first issue, we propose to impose cross supervision based on a multi-input multi-output (MIMO) model rather than multiple independent models. MIMO is an efficient method to resolve the problem of expensive costs during ensembling. 
Through MIMO, multiple predictions can be obtained under a single forward pass, and then the purpose can be achieved almost ``free'' \cite{havasi2020training}.
Therefore, it is natural expending the MIMO to cross supervision to reduce the training cost.
In our USCS, instead of ensemble models, cross supervision is realized by one MIMO model, which is called self cross supervision in this work.

For the second issue, we propose employing uncertainty guided the process of learning with wrong pseudo labels. Uncertainty is used to evaluate the quality of predictions without ground truth. Generally, regions with large uncertainty represent poor prediction and vice versa \cite{abdar2021review}.
For the task at hand, uncertainty can be used as the guided information to indicate the confidence of the pseudo labels of unlabelled samples and supervise the cross supervision process by reducing the effects of wrong pseudo labeling, and such proposed method is called uncertainty-guided learning method.

In conclusion, our contributions are:
\begin{enumerate}
	\item  We firstly propose a self cross supervision method with a multi-input multi-output (MIMO) model. The MIMO model is specially designed for the semantic segmentation task. Our method realizes cross supervision through enforcing the consistency between the outputs of MIMO, and greatly reduces the training cost of the model.
	\item We propose uncertainty-guided learning for semi-supervised semantic segmentation to improve the performance of the model, which uses the uncertainty information as the confidence of the pseudo labels and supervises the learning process by reducing the effects of wrong pseudo labeling. 
	\item Experiments demonstrate that our proposed model surpasses most of the current state-of-the-art methods. Moreover, compared with cross supervision, our method can achieve competitive performance while greatly reducing training costs.
\end{enumerate}

\section{Related work}

\subsection{Semantic segmentation}
Semantic segmentation is a pixel-wise classification task, which marks each pixel of the image with the corresponding class.
Most of the current semantic segmentation models are based on the encoder-decoder structure \cite{badrinarayanan2017segnet,noh2015learning,ronneberger2015u}. The encoder reduces the spatial resolution generating a high-level feature map, and the decoder gradually restores spatial dimension and details. Fully convolutional neural networks (FCN) \cite{long2015fully} is the first encoder-decoder-based segmentation model. The subsequent works improve the context dependence by dilated convolutions \cite{yu2015multi,chen2017deeplab}, maintaining high resolution \cite{sun2019deep,wang2020deep}, pyramid pooling \cite{zhao2017pyramid,yang2018denseaspp}, and self-attention mechanism \cite{vaswani2017attention}.
DeepLabv3+ \cite{chen2018encoder} is one of the state-of-the-art methods, which is employed as the segmentation model in this work.

\subsection{Semi-supervised learning}
Semi-supervised learning focuses on high performance using abundant unlabeled data under limited labeled data, so as to alleviate the training dependence on labels \cite{lee2013pseudo,kipf2016semi,zhu2005semi}.
Most of the current semi-supervised learning methods are based on empirical assumptions of the image itself, such as smoothness assumption, and low-density assumption \cite{van2020survey}. 

Based on the smoothness assumption, prior works use the consistent regularization semi-supervised method, which encourage the model to predict the similar output for the perturbed input. This kind of works tries to minimize the difference between perturbed samples generated by data augmentations, e.g., Mean Teacher \cite{tarvainen2017mean}, VAT \cite{miyato2018virtual} and UDA \cite{xie2020unsupervised}. As for the low-density assumption, the pseudo label based semi-supervised learning \cite{lee2013pseudo,pham2021meta,xie2020unsupervised} is the representative method, which realizes the low-density separation by minimizing the conditional entropy of class probability for the unlabeled data. In order to utilize the merits of different assumptions, prior works also propose effective methods based on both or more. Among these methods, joint learning with the pseudo label and consistent regularization is a successful one and has achieved impressive performance, such as MixMatch \cite{berthelot2019mixmatch}, FixMatch \cite{sohn2020fixmatch} and DivideMix \cite{li2020dividemix}. Our approach utilizes consistent regularization and the pseudo label to construct semi-supervised learning.

\subsection{Semi-supervised semantic segmentation}
As a dense prediction task, semantic segmentation is laborious and time-consuming in manual annotations. Therefore, using unlabeled images to improve model performance is an effective way for cost reduction. Most of the semi-supervised semantic segmentation approaches are based on the consistent regularization \cite{olsson2021classmix,french2019semi,zou2020pseudoseg,lai2021semi}.For example, PseudoSeg \cite{zou2020pseudoseg} enforces the consistency of the predictions with weak and strong data augmentations, similar to FixMatch \cite{sohn2020fixmatch}. CAC \cite{lai2021semi} utilizes contextual information to maintain the consistency between features of the same identity under different environments. CCT \cite{ouali2020semi} maintains the agreement between the predictions from the features with various perturbations.
GCT \cite{ke2020guided} and CPS \cite{chen2021semi} adopt different model structures or model initializations to generate the perturbations of predictions and achieve state-of-the-art performance. However, the training cost of time and memory for ensemble models is expensive in GCT and CPS. Different from prior works, our approach enforces the consistency of predictions from a multi-input multi-output network and greatly reduces the training costs.

\section{Method}

In the following sections, we first introduce the overview of USCS in Sec.~\ref{section:3.1}. The self cross supervision with MIMO model is proposed in Sec.~\ref{section:3.2}. To ameliorate pseudo label quality, we propose the uncertainty-guided learning in Sec.~\ref{section:3.3}.

As a common semi-supervised learning task, a dataset $\mathcal{X}$ consisting of labeled images $\mathcal{X}_l$ with labels $\mathcal{Y}$ and unlabeled images $\mathcal{X}_{ul}$ is employed to train a segmentation network. In our USCS, we extra applied transformation $T$ on unlabeled images $\mathcal{X}_{ul}$ got the transformed images $\mathcal{X}^T_{ul}=T(\mathcal{X}_{ul})$. Both unlabeled images $\mathcal{X}_{ul}$ and transformed images $\mathcal{X}^T_{ul}$ are employed to construct self cross supervision.

\begin{figure}
	\centering
	\includegraphics[width=\linewidth]{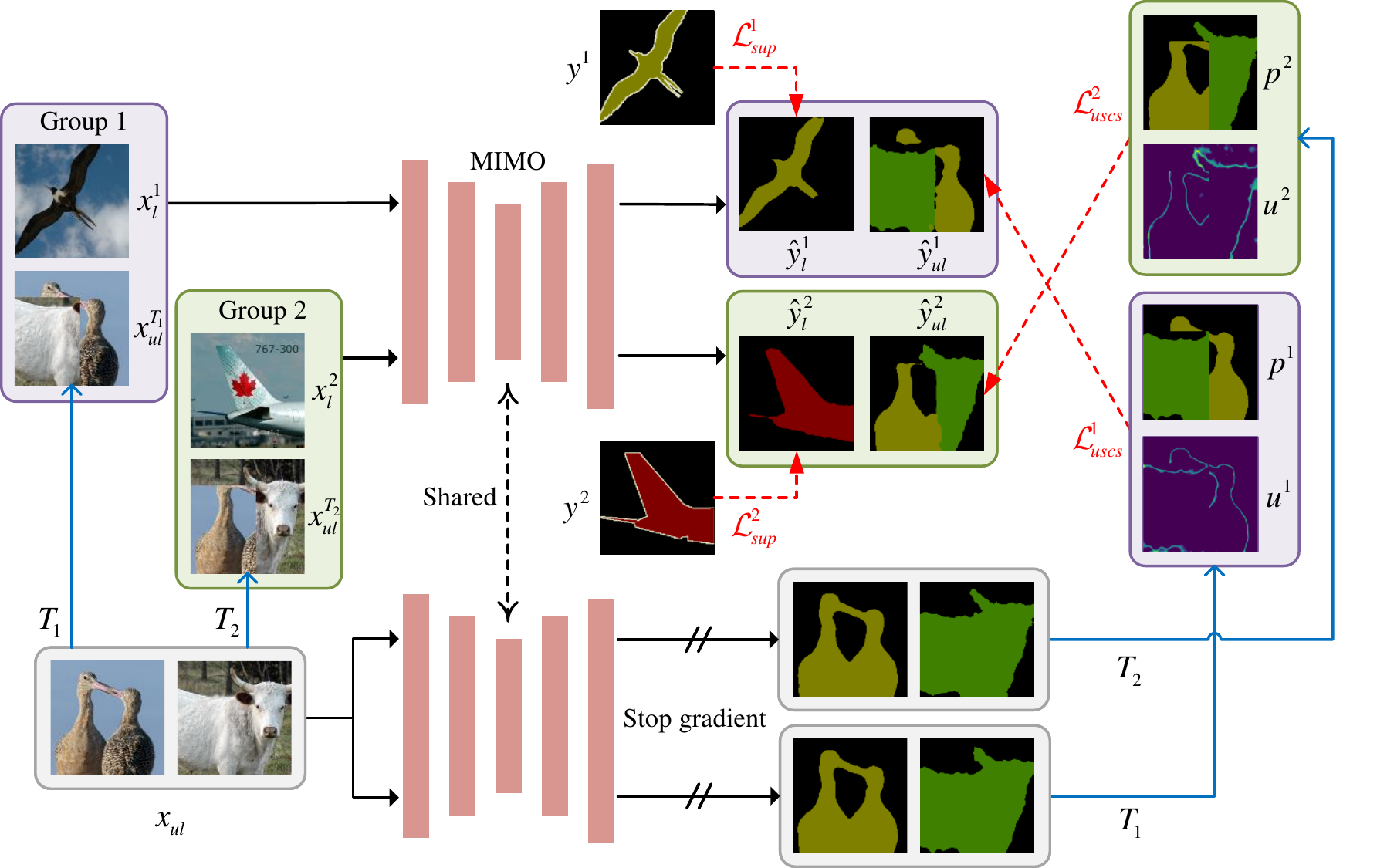}
	\caption{\textbf{The USCS Framework}. We aim to maintain the consistency between the predictions from a multi-input multi-output (MIMO) model. Since MIMO accepts two different group images, we adopted transformation consistency to realize the purpose.}
	\label{fig:2}
\end{figure}

\subsection{Overview of USCS}{\label{section:3.1}}
The USCS framework is shown in Fig.~\ref{fig:2}. In contrast to the general cross supervision method using several independent models, we instead employ a multi-input multi-output (MIMO) model. Speciﬁcally, the MIMO model $F$ has two input and output branches which can be seen as the subnetworks with shared parameters, accepting two groups independently sampled data $G_k\left( k\in \left\{ 1,2 \right\} \right)$ and output corresponding segmentation results. 
In USCS, each group data is denoted as $G_k=\left\{{x}^{k}_l,x^{T_k}_{ul}\right\}$ ($ {x}^{k}_l\in{\mathcal{X}_l}, x^{T_k}_{ul}\in{\mathcal{X}^T_{ul}} $). For $k \in \left\{1,2\right\}$, $x^1_l$ and $x^2_l$ are the labeled images with different batch sampling order, $x^{T_1}_{ul}$ and $x^{T_2}_{ul}$ are the transformed images with distinct transformation $T_k$.

Given an image $x^k \in {G}_k$, the MIMO model $F$ first predicts $\hat{y}^k=\{\hat{y}^k_l,\hat{y}^k_{ul}\}$, where $\hat{y}^k_l=F(x^k_l)$ and $\hat{y}^k_{ul}=F(x^{T_k}_{ul})$. As common semantic segmentation models, the prediction $\hat{y}^k_l$ is supervised by its corresponding ground-truth $y\in \mathcal{Y}$ as:
\begin{align}
	{\mathcal{L}^k_{sup}}\left( {x^k_{l}},y \right)=\frac{1}{\left| \Omega  \right|}\sum\limits_{i\in \Omega }{{{\ell }_{ce}}(\hat{y}^k_l(i),y(i))},
\end{align}
where $\ell_{ce}(*)$ is the standard Cross Entropy loss, and $\Omega$ is the region of image with size $H\times W$.

To explore the unlabeled images, we repeat the original unlabeled images $x_{ul}$ twice, the MIMO model $F$ makes two groups independent predictions ${F}^{1}({{x}_{ul}})$ and ${F}^{2}({{x}_{ul}})$ on the same images $x_{ul}$ as shown at the bottom of Fig.~\ref{fig:2}. Then the same transformation $T_1$ and $T_2$ are respectively performing on ${F}^{2}({{x}_{ul}})$ and ${F}^{1}({{x}_{ul}})$, obtaining $p^1=T_1({F}^{2}({{x}_{ul}}))$, $p^2=T_2({F}^{1}({{x}_{ul}}))$. Besides, the uncertainties $u^1$ and $u^2$ are estimated for two transformed predictions $p^1$ and $p^2$, respectively. Then, $p^1$ guided by the uncertainty $u^1$ is regarded as the pseudo labels of $x_{ul}^{T_1}$ to supervise $\hat{y}^1_{ul}$. Similarly, the same operation is used to supervise $\hat{y}^2_{ul}$ based on $p^2$ and $u^2$.
We call the above process uncertainty-guided self cross supervision. The constraint $\mathcal{L}_{uscs}$ and more details are described in Sec.~\ref{section:3.2} and Sec.~\ref{section:3.3}.

Finally, our method for the training of MIMO model $F$ joint the two constraints on both the labeled and unlabeled images which can be written as:
\begin{align}
	{\mathcal{L}}\left( \mathcal{X},\mathcal{Y} \right)=\sum\limits_{k=1,2}(\frac{1}{\left| \mathcal{X}_l \right|}\sum\limits_{x^k_l\in \mathcal{X}_l }{\mathcal{L}^k_{sup}}\left( {x^k_{l}},y \right)+\frac{1}{\left| \mathcal{X}_{ul} \right|}\sum\limits_{x_{ul}\in \mathcal{X}_{ul} }\lambda \mathcal{L}_{uscs}^{k}({{x}_{ul}})),
\end{align}
where $\lambda$ is the trade-off weight to balance the USCS constraint.

\subsection{Self Cross Supervision with MIMO Model}{\label{section:3.2}}
The proposed self cross supervision is implemented over the MIMO model. Before presenting self cross supervision, the MIMO model used in USCL is firstly introduced. Based on the fact that neural networks are heavily overparameterized models \cite{havasi2020training}, we can train a MIMO model containing multiple independent subnetworks and acquire multiple predictions of one input under a single forward pass of the model. 
Different from the single neural network archtiecture, the MIMO model replaces the single input layer by $N$ input layers, which can receive $N$ datapoint as inputs. And $N$ output layers are added to make $N$ predictions based on the feature before output layers. 
Compared with a single model, the MIMO model obtains the performance of ensembling with the cost of only a few increased parameters and calculations.
\begin{figure}
	\centering
	\includegraphics[width=\linewidth]{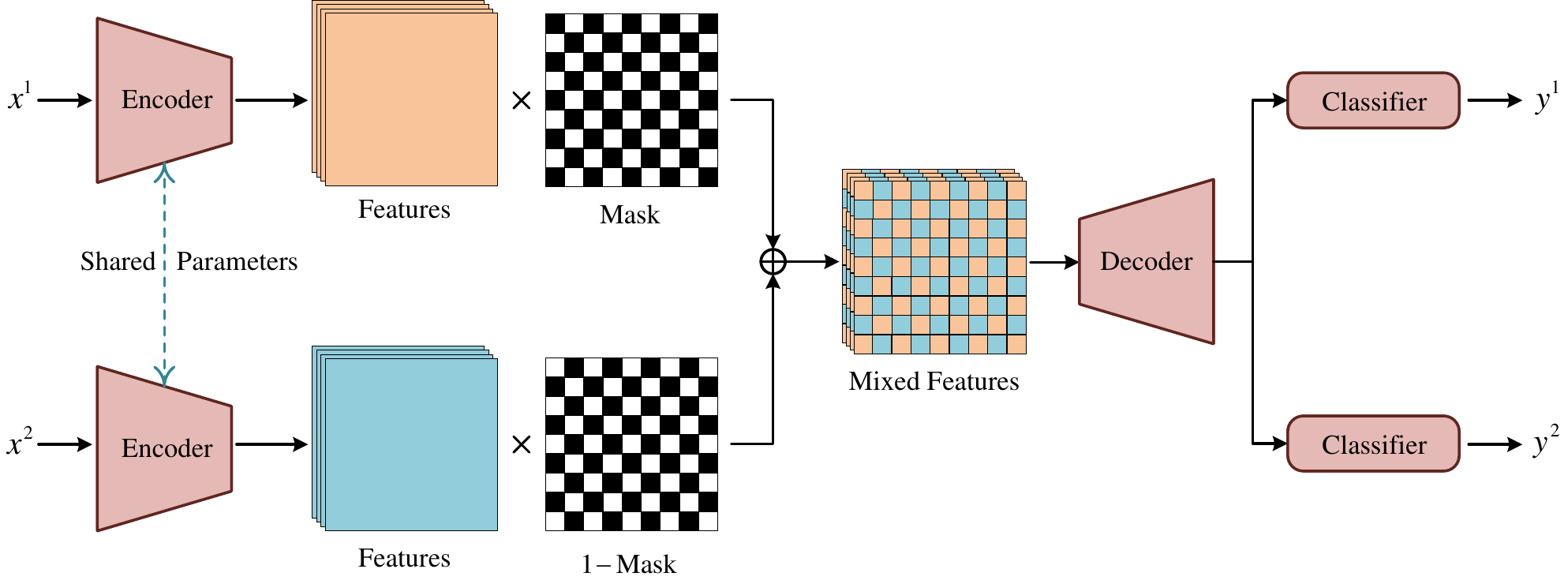}
	\caption{\textbf{The structure of MIMO segmentation model}. The features after the encoder are fused by the grid mix.}
	\label{fig:3}
\end{figure}

In USCS, we construct a MIMO model with two inputs and outputs, whose structure is shown in Fig.~\ref{fig:3}. 
For better extract object features, the entire encoder part is utilized as the input layer of the model (the original MIMO model employs the first convolutions layers of the model as the input layer). However, two independent encoders (the input layer) increase the model parameters and computation. We share the parameters of two encoders to avoid this problem. The features of two inputs extracted by the encoder must be fused before entering the decoder. 
To effective combines inputs into a shared representation, the grid mix is adopted to replace the original summing method \cite{havasi2020training} in MIMO as:
\begin{align}
	{{\mathcal M}_{gridmix}}({{f}_{1}},{{f}_{2}})={{\mathbbm{1}}_{\mathcal{M}}}\odot {{f}_{1}}+(\mathbbm{1}-{{\mathbbm{1}}_{\mathcal{M}}})\odot {{f}_{2}},
\end{align}
where ${f}_{1}$ and ${f}_{2}$ are the features of two inputs, respectively; 
${\mathbbm{1}}_{\mathcal{M}}$ is a binary grid mask with grid size $g$. 

The self cross supervision enforces two predictions of MIMO learn from each other. The output $y^1$ is considered the pseudo label to supervise the output $y^2$, vice versa. As mentioned previously, two inputs of MIMO are different, while the self cross supervision is feasible only when the inputs are the same. We overcome this issue by introducing the transformation consistency regularization \cite{mustafa2020transformation}, which assumes that the prediction $F(T(x))$ of the transformed image $T(x)$ must be equal to the transformed prediction $T(F(x))$ of the original image $x$. 

As shown in Fig.~\ref{fig:2}, the MIMO model $F$ predicts two transformed unlabeled images $x^{T_1}_{ul}$ and $x^{T_2}_{ul}$, obtaining $\hat{y}^1_{ul}$ and $\hat{y}^2_{ul}$. Self cross supervision expects two outputs of the MIMO model to supervise each other. However, the semantics of the outputs $\hat{y}^1_{ul}$ and $\hat{y}^2_{ul}$ are different. To achieve the self cross supervision, we input the original unlabeled image $x_{ul}$ to the MIMO model, getting two individual predictions $F^1(x_{ul})$ and $F^2(x_{ul})$ without gradient. We further obtain two transformed predictions $p^1=T_1(F^{2}(x_{ul}))$ and $p^2=T_2(F^{1}(x_{ul}))$ by performing the transformation $T_1$ and $T_2$, respectively. The transformed predictions $p^1$ should have the similar semantics with $\hat{y}^1_{ul}$, thus we regard $p^1$ as the pseudo label of $\hat{y}^1_{ul}$. Similarly, the transformed prediction $p^2$ is considered as the pseudo label to supervise $\hat{y}^2_{ul}$.

Through the above process, the MIMO model $F$ can realize cross supervision by itself. The self cross supervision constraint on unlabeled data is defined as:
\begin{align}\label{eq:scl}
	\mathcal{L}_{scs}^{k}({{x}_{ul}})=\frac{1}{\left| \Omega  \right|}\sum\limits_{i\in \Omega }{{{\ell }_{ce}}(\hat{y}^k_{ul}(i),p^k(i))}.
\end{align}

\subsection{Uncertainty-guided Learning}{\label{section:3.3}}
The pseudo label obtained from the prediction exits noise, especially in the early stage during the training, where the model with poor performance produces plenty of inaccurate pseudo labels. The noisy pseudo label will mislead the model and interfere with the optimization direction in cross supervision. In addition, the noise caused by one model is likely to propagate to another model through cross supervision, resulting in the accumulation and propagation of errors and hindering the performance. 
It is necessary to filter the pseudo label with inferior quality to improve the overall performance of the model. 

\begin{figure}
	\centering
	\includegraphics[width=0.7\linewidth]{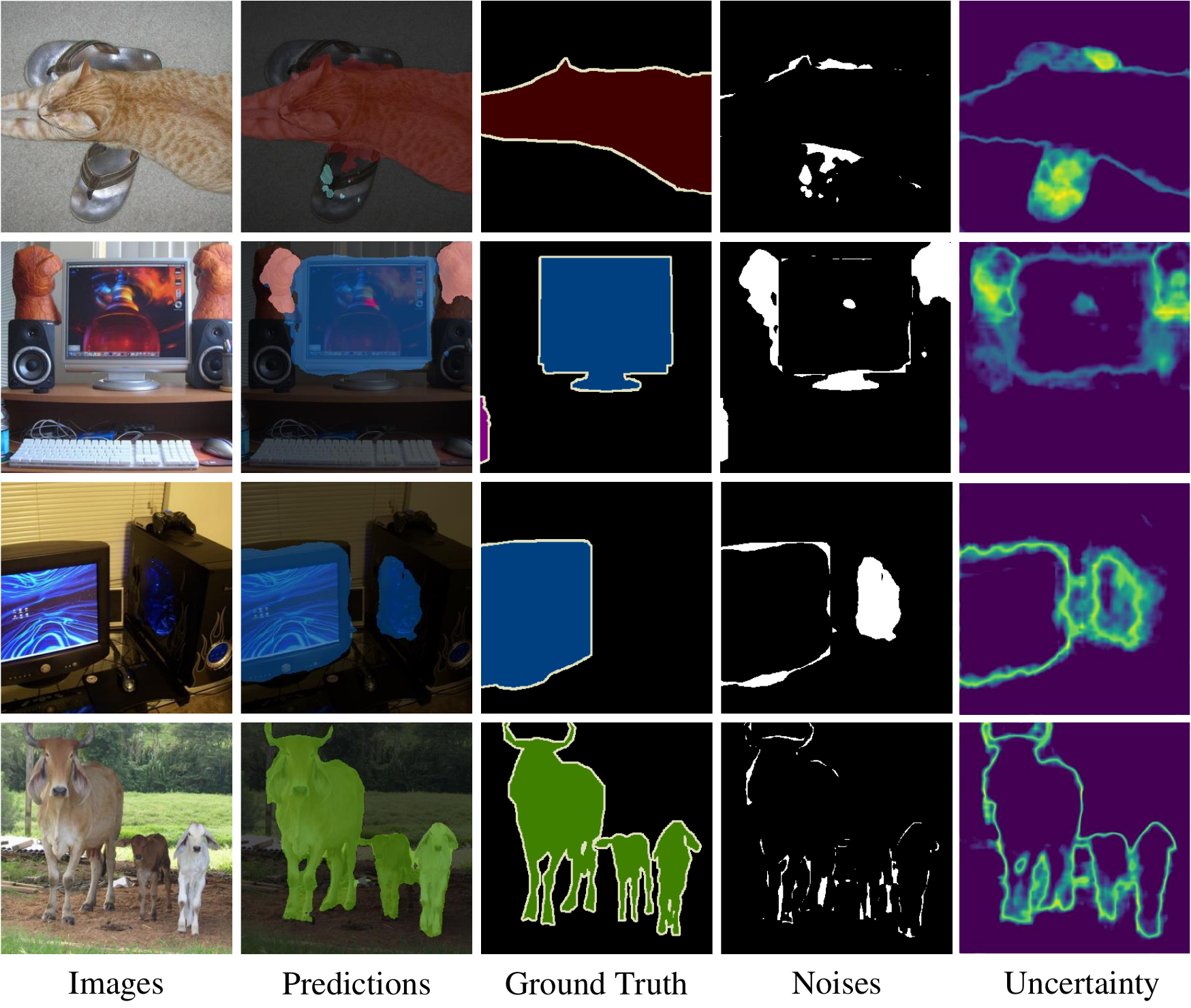}
	\caption{\textbf{Uncertainty visualization}. Highly bright regions represent large uncertainties in the  uncertainty map.}
	\label{fig:4}
\end{figure}

Uncertainty estimation is an effective method to evaluate noise in prediction \cite{lakshminarayanan2017simple}. Noise often exists in regions with large uncertainties. Fig.~\ref{fig:4} shows the uncertainty visualization. 
Based on this observation, we propose to employ uncertainty to guide the pseudo label with noise in cross supervision. Firstly, we estimate the uncertainty of pseudo label through the Shannon Entropy \cite{shannon2001mathematical}, which is defined as:
\begin{align}
	U=-\sum\limits_{c=1}^{C}{p(c)\log p(c)},
\end{align}
where $p$ is the softmax predicted vector with $C$ channel. We normalize $U$ into range $(0,1)$, and set $\widehat{U}=1-U$. Then, the pseudo label can be divided into confident and uncertain regions by setting a threshold $\gamma$. We fully receive the pixels in the confident region, which are regarded as the true label. As for the uncertain regions, we assign low loss weights to high uncertain pixels. Thus the model can also learn from the pixels in the uncertain regions, which avoids the loss of useful information.
We define the uncertainty weight mask as:
\begin{align}\label{equ:7}
	W=\left\{
	\begin{array}{rcl}
		1    &   ~~~~& {\widehat{U}\ge \gamma}  \\ 
		{\widehat{U}}/{\gamma }  &   & {\widehat{U}<\gamma}  \\ 
	\end{array} \right.
\end{align}

In the end, we multiply the weight mask $W$ to the self cross supervision constraint and rewrite the Eq.~\ref{eq:scl}, getting the uncertainty-guided self cross supervision constraint:
\begin{align}
	\mathcal{L}_{uscs}^{k}({{x}_{ul}})=\frac{1}{\left| \Omega  \right|\left| sum(W)  \right|}\sum\limits_{i\in \Omega }{W(i)\cdot {{\ell }_{ce}}(\hat{y}^k_{ul}(i),p^k(i))},
\end{align}
where $sum(*)$ means the sum of elements in the matrix.

\section{Expermients}
%In the following sections, we show the experimental setup in Sec.~\ref{section:4.1}. The comparison with supervised baselines and state-of-the-art approaches under different partition protocols is presented in Sec.~\ref{section:4.2}. The ablation study is performed in Sec.~\ref{section:4.3}.
\subsection{Experimental Setup}\label{section:4.1}
\subsubsection{Datasets.}
PASCAL VOC 2012 \cite{everingham2015pascal} is the most prevalent benchmark for semi-supervised semantic segmentation with 20 object classes and one background class. The standard dataset contains 1464 images for training, 1449 for validation, and 1456 for testing. 
Following previous works \cite{chen2021semi}, we adopt the augmented set provided from SBD \cite{hariharan2011semantic} as our entire training set, which contains 10582 images.

\subsubsection{Implementation details.}
The results are obtained by training the MIMO model, modified on the basis of Deeplabv3+ \cite{chen2018encoder}.
We regard the backbone of the segmentation model as the encoder, whose weights are initialized with the pre-trained model on ImageNet \cite{mmseg2020}. 
The other components except the final classifier are considerd as the decoder which are initialized randomly. 

Following the previous works \cite{chen2021semi}, we utilize ``poly" learning rate decay policy where the base learning rate is scaled by ${{(1-{iter}/{max\_iter}\;)}^{0.9}}$. 
Mini-batch SGD optimizer is adopted with the momentum and weight decay set to 0.9 and $10^{-4}$ respectively. 
During the training, images are randomly cropped to $320\times320$, random horizontal flipping with a probability of 0.5, and random scaling with a ratio from 0.5 to 2.0 are adopted as data augmentation. We train PASCAL VOC 2012 for $3\times10^{4}$ iters with batch size set to 16 for both labeled and unlabeled images. The base learning rates are 0.001 for backbone parameters and 0.01 for others. The trade-off weight $\lambda$ is set to 1 after adjustment.

Besides, we found that the MIMO model based on Deeplabv3+ cannot accommodate two independent subnetworks due to the limited capacity. Thus, we relax independence same as \cite{havasi2020training} by sampling two same inputs from the training set with probability $\rho$, i.e., the input $x_2$ of the MIMO model is set to be equal to $x_1$ with probability $\rho$. During the training, we employ CutMix \cite{yun2019cutmix} as transformation, same as \cite{chen2021semi}. We average two outputs of the MIMO model to generate the final results for evaluation.

\subsubsection{Evaluation.}
We use the mean Intersection-over-Union (mIoU) as the evaluation metric as a common practice. To evaluate training time and memory cost reduction in USCS, Multiply–Accumulate Operations (MACs) and the number of parameters are adopted as the metric. Besides, we employ the non overlap ratio for the outputs of the MIMO model as metric to measure the diversity of subnetworks. The low non overlap ratio means poor diversity.

\subsection{Results}\label{section:4.2}
In this section, we report the results compared with supervised baselines and other SOTA methods in different partition protocols, i.e., the full training set is split with 1/16, 1/8, 1/4, and 1/2 ratios for labeled images and the remainder as unlabeled images.
\begin{figure}[htbp]
	\centering
	\includegraphics[width=0.8\linewidth]{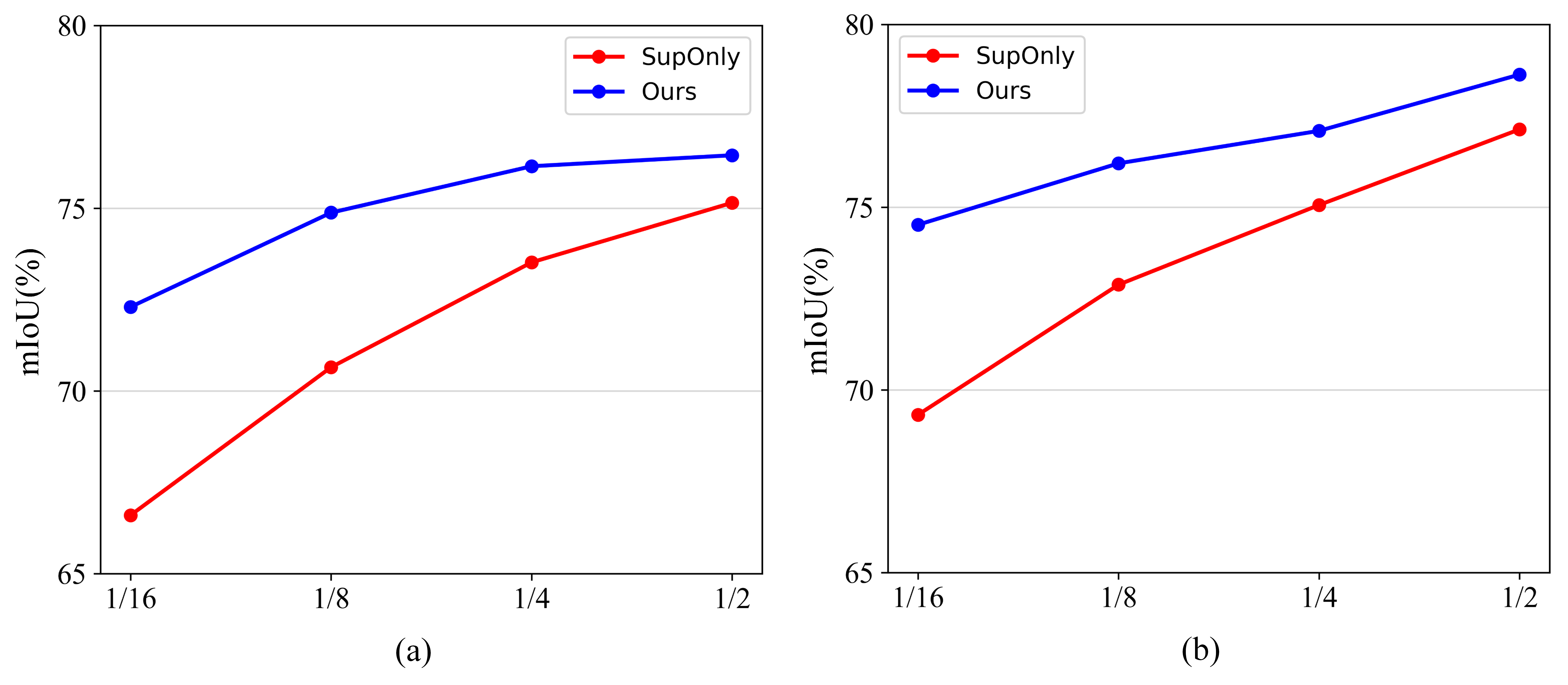}
	\caption{\textbf{Improvements over the supervised baseline} on PACAL VOC 2012 with (a) Resnet-50 and (b) Resnet-101}
	\label{fig:5}
\end{figure}

\subsubsection{Improvements over Supervised Baselines.}
Fig.~\ref{fig:5} illustrates the improvements of our approach compared with full supervised learning (trained with the same partition protocol). Speciﬁcally, our method outperforms the supervised baseline by 5.70\%, 4.23\%, 2.63\%, and 1.30\% under 1/16, 1/8, 1/4, and 1/2 partition protocols separately with Resnet-50. On the other settings, the gains obtained by our approach are also stably: 5.20\%,3.32\%, 2.03\%, and 1.50\% under 1/16, 1/8, 1/4, and 1/2 partition protocols separately with Resnet-101.
\setlength{\tabcolsep}{5pt}
\begin{table}[htbp]
	\centering
	\caption{\textbf{Comparison with SOTA} on PASCAL VOC 2012. All the approachs are based on Deeplabv3+. The $\ast$ indicates the approaches re-implemented by \cite{chen2021semi}. Best results are in bold; suboptimal results are in italics.}
	\begin{tabular}{ccccc|cccc}
		\hline\noalign{\smallskip}
		\multicolumn{1}{c}{\multirow{2}{*}{\textbf{Methods}}} & \multicolumn{4}{c|}{Resnet-50} & \multicolumn{4}{c}{Resnet-101}  \\
		
		%\cline{2-9}
		
		& 1/16 & 1/8  & 1/4  & 1/2  & 1/16 & 1/8 & 1/4  & 1/2 \\
		\noalign{\smallskip}
		\hline\noalign{\smallskip}
		MT$\ast$\cite{tarvainen2017mean}    & 66.77 & 70.78 & 73.22 & 75.41 & 70.59 & 73.20  & 76.62 & 77.61 \\
		CutMix-Seg$\ast$\cite{french2019semi} & 68.90  & 70.70  & 72.46 & 74.49 & 72.56 & 72.69 & 74.25 & 75.89 \\
		CCT$\ast$\cite{ouali2020semi}   & 65.22 & 70.87 & 73.43 & 74.75 & 67.94 & 73.00    & 76.17 & 77.56 \\
		GCT$\ast$\cite{ke2020guided}   & 64.05 & 70.47 & 73.45 & 75.20  & 69.77 & 73.30  & 75.25 & 77.14 \\
		CAC\cite{lai2021semi}   & 70.10  & 72.40  & 74.00   & - & 72.40  & 74.60  & 76.30  &- \\
		CPS\cite{chen2021semi} & \emph{71.98} & \emph{73.67} & \emph{74.90}  & \emph{76.15} & \emph{74.48} & \textbf{76.44} & \textbf{77.68} & \textbf{78.64} \\
		\noalign{\smallskip}
		\hline
		\noalign{\smallskip}
		Ours & \textbf{72.30}  & \textbf{74.88} & \textbf{76.15} & \textbf{76.45} & \textbf{74.52} & \emph{76.20} & \emph{77.09} & \emph{78.63} \\
		\noalign{\smallskip}
		\hline
	\end{tabular}
	\label{tab:1}
\end{table}
\subsubsection{Comparison with SOTA.}
The results compared with other semi-supervised approaches are shown in Tab.~\ref{tab:1}. Our method performs better than most methods under different partition protocols with Resnet-50 and Resnet-101 as backbones. Compared with CAC \cite{lai2021semi}, our approach improves by 2.2\%, 2.48\%, 2.15\% under 1/16, 1/8, and 1/4 partition protocols separately with Resnet-50. 
Compared with CPS \cite{chen2021semi}, the advantage of our method is a great reduction in the number of parameters and calculations as shown in Tab.~\ref{tab:2}. We acquired 40.5\% and 49\% economization on MACs and parameters with Resnet-50, which signify the cost decrease of training time and memory. Besides, our method only needs twice forward pass, while CPS needs four times.
As for accuracy, our method achieves around 1\% improvement in all cases with Resnet-50. In the cases with Resnet-101, the performance of our approach declined slightly except for 1/16 partitions. We think the main reason for the caused decline is that insufficient capacity of the MIMO model limits the diversity of two subnetworks and further influences the performance of self cross supervision.

\setlength{\tabcolsep}{6pt}
\begin{table}[htbp]
	\centering
	\caption{\textbf{Training cost comparison} with CPS \cite{chen2021semi} and SupOnly.}
	\begin{tabular}{ccc|cc|c}
		\hline
		\noalign{\smallskip}
		\multicolumn{1}{c}{\multirow{2}[0]{*}{\textbf{Methods}}} & \multicolumn{2}{c|}{Resnet-50} & \multicolumn{2}{c|}{Resnet-101} & \multicolumn{1}{c}{\multirow{2}[0]{*}{Forward passes}} \\
		& \multicolumn{1}{c}{MACs(G)} & \multicolumn{1}{c|}{Params(M)} & \multicolumn{1}{c}{MACs(G)} & \multicolumn{1}{c|}{Params(M)} &  \\
		\noalign{\smallskip}
		\hline
		\noalign{\smallskip}
		SupOnly & 23.84 & 39.78 & 31.45 & 58.77 & 1      \\
		CPS   & 95.36 & 79.56 & 125.80  & 117.54 & 4      \\
		Ours  & 56.74 & 40.49 & 71.94 & 59.48 & 2       \\
		\noalign{\smallskip}
		\hline
	\end{tabular}
	\label{tab:2}
\end{table}

\subsection{Ablation Study}\label{section:4.3}
This section conducts the ablation study to exhibit the roles of self cross supervision (SCS) and uncertainty-guided learning (UL) in our method. Besides, the influences of uncertainty threshold $\gamma$, feature fusion methods, and input repetition probability $\rho$ are reported, respectively. All the experiments are run based on 1/8 partition protocols on PASCAL VOC 2012.
\begin{figure}[htbp]
	\centering
	\includegraphics[width=\linewidth]{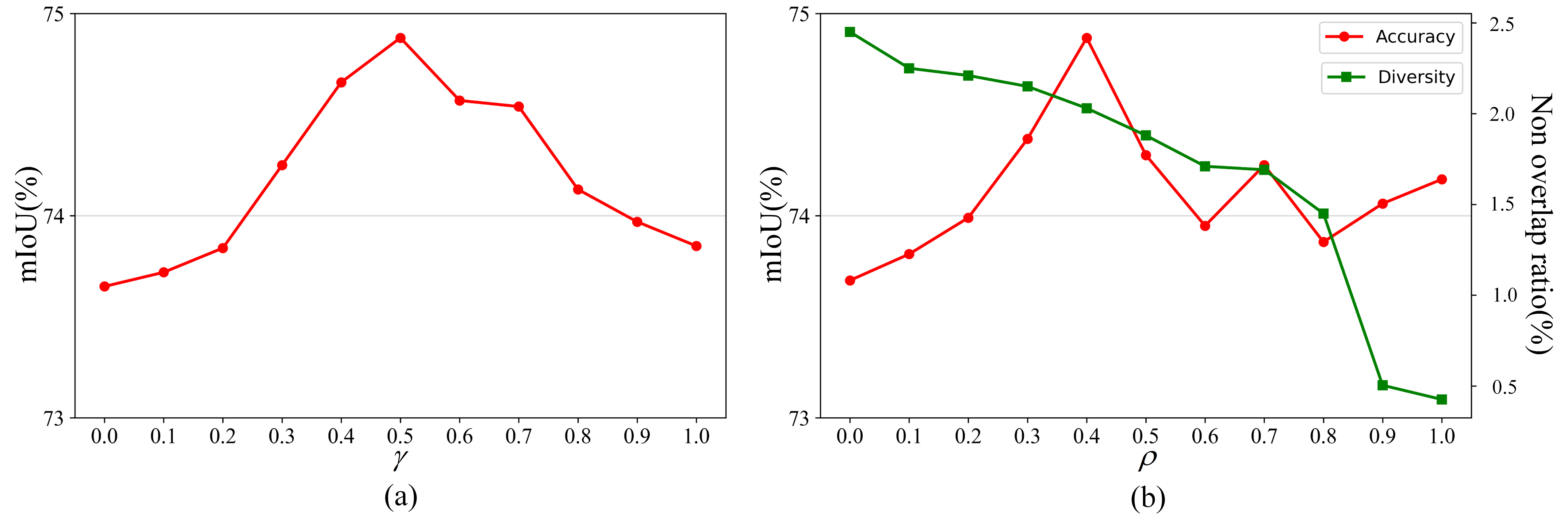}
	\caption{\textbf{The ablation study} on (a) uncertainty threshold $\gamma$ and (b) input repetition probability $\rho$. }
	\label{fig:6}
\end{figure}

\subsubsection{Uncertainty guided self cross supervision.}
The contribution of self cross supervision and uncertainty-guided learning are shown in Tab.~\ref{tab:3}.
It is important to note that we adopt the result of CPS with CutMix augmentation\cite{chen2021semi} as a baseline to ensure fairness. We report a slight decline in performance after replacing CPS with SCS. 
While, the improvements yielded by UL are 1.23\% with the Resnet-50. We can see that SCS heavily reduces training costs of time and memory, and UL improves the performance without extra cost.

\setlength{\tabcolsep}{8pt}
\begin{table}[htbp]
	\centering
	\caption{\textbf{Ablation study of different components} under 1/8 partition protocols on PASCAL VOC 2012.}
	\begin{tabular}{ccc|ccc}
		\hline
		\noalign{\smallskip}
		\multicolumn{1}{c}{\multirow{2}[0]{*}{CPS}} & \multicolumn{1}{c}{\multirow{2}[0]{*}{SCS}} & \multicolumn{1}{c}{\multirow{2}[0]{*}{UL}} & \multicolumn{3}{|c}{Resnet-50}  \\
		&       &       & \multicolumn{1}{c}{mIoU(\%)$\uparrow$} & \multicolumn{1}{c}{MACs(G)$\downarrow$} & \multicolumn{1}{c}{Params(M)$\downarrow$}  \\
		\noalign{\smallskip}
		\hline
		\noalign{\smallskip}
		\checkmark  &   &   & 73.67 & 95.36 & 79.56\\
		& \checkmark  &   & 73.65 &  56.74 & 40.49 \\
		& \checkmark     & \checkmark     & \textbf{74.88} & \textbf{56.74} & \textbf{40.49} \\
		\noalign{\smallskip}
		\hline
		\label{tab:3}
	\end{tabular}
\end{table}% 

\subsubsection{Uncertainty threshold $\gamma$.}
\begin{figure}[htbp]
	\centering
	\includegraphics[width=0.6\linewidth]{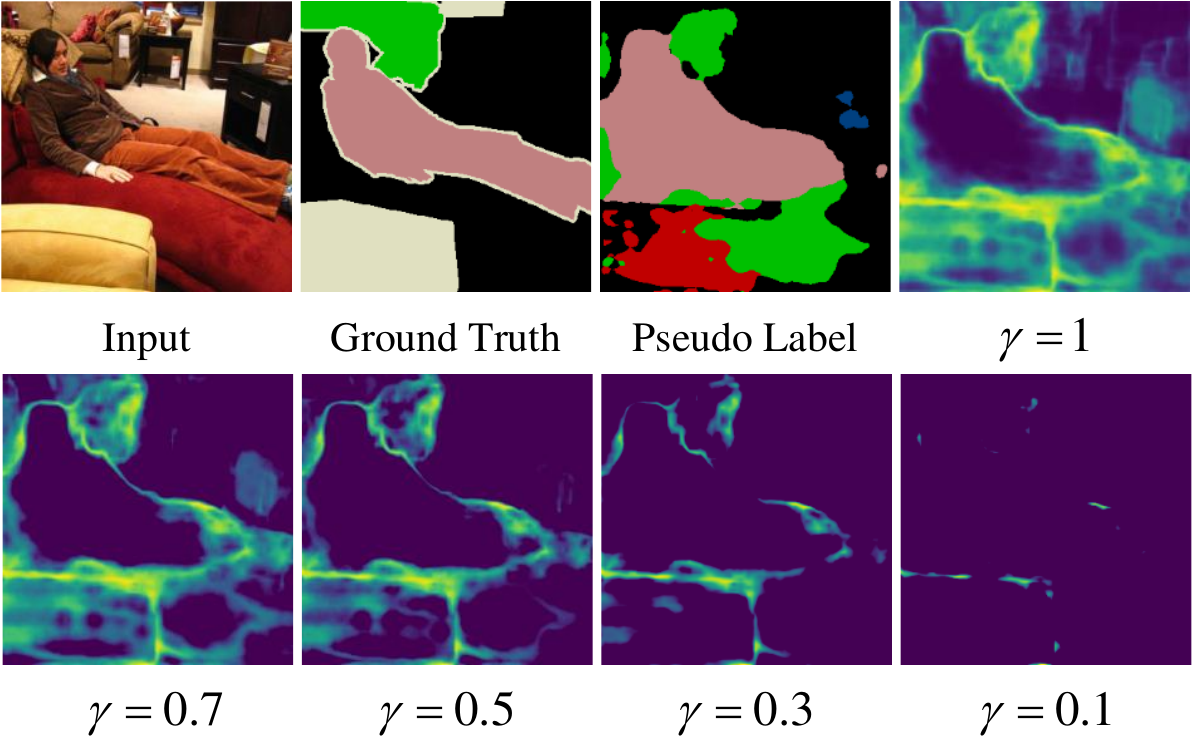}
	\caption{\textbf{Visual comparison with different uncertainty threshold $\gamma$}.}
	\label{fig:7}
\end{figure}

We investigate the influence of threshold $\gamma$ used to control the uncertain weight mask as shown in Equation.~\ref{equ:7}. The results in Fig.~\ref{fig:6}(a) show that: with the increase of $\gamma$, the model reduces the weight of learning for noisy pixels in pseudo label and performs best when $\gamma=0.5$. When the continuous increase of $\gamma$, the performance degrades due to the model tends to regard all pixels in pseudo label as noise, reducing the weight of confident pixels in pseudo label. We visualize the effect of threshold $\gamma$ to uncertainty in Fig.~\ref{fig:7}. 

\subsubsection{Input repetition probability $\rho$.}
We show the influence of probability $\rho$ on both accuracy and diversity in Fig.~\ref{fig:6}(b). When $\rho=0$, the training images are sampled independently for both subnetworks, and the MIMO model acquired great diversity but poor accuracy as it can not contain two independent subnetworks. As $\rho$ grew, the diversity of the MIMO model gradually decayed, the independence of the subnetwork is relaxed to release the limited model capacity. The performance reaches the peak at $\rho=0.4$, where get a trade-off between the diversity and the capacity of the MIMO model. 
\setlength{\tabcolsep}{6pt}
\begin{table}[htbp]
	\centering
	\caption{\textbf{The affects of feature fusion methods} on mIoU(\%) and non overlap ratio(\%).}
	\begin{tabular}{cccccc}
		\hline
		\noalign{\smallskip}
		\multicolumn{1}{c}{\multirow{2}[0]{*}{{Fusion methods}}} & \multicolumn{4}{c}{Grid mix}  & \multicolumn{1}{c}{\multirow{2}[0]{*}{Summing}} \\
		\noalign{\smallskip}
		\cline{2-5}
		\noalign{\smallskip}
		& 1     & 3     & 5     & 7     &  \\
		\noalign{\smallskip}
		\hline
		\noalign{\smallskip}
		mIoU(\%)  & 74.88 & 74.10  & 73.45 & 73.64 & 73.32 \\
		Non overlap ratio(\%) & 2.03 & 2.53 & 1.64 & 1.40 & 0.94 \\
		\noalign{\smallskip}
		\hline
	\end{tabular}%
	\label{tab:4}
\end{table}

\subsubsection{Feature fusion.}
We show the influence of feature fusion methods, summing and grid mix, on both accuracy and diversity in Tab.~\ref{tab:4}.
The block size $g$ of the grid mix is set as 1, 3, 5, and 7. We can see that the grid mix surpasses the summing feature fusion method on both mIoU scores and non overlap ratios. The accuracy of the MIMO model decreases as $g$ increases, while the diversity reaches the top at $g=3$. We use $g=1$ in our method for all the experiments.

\begin{figure}[htbp]
	\centering
	\includegraphics[width=0.6\linewidth]{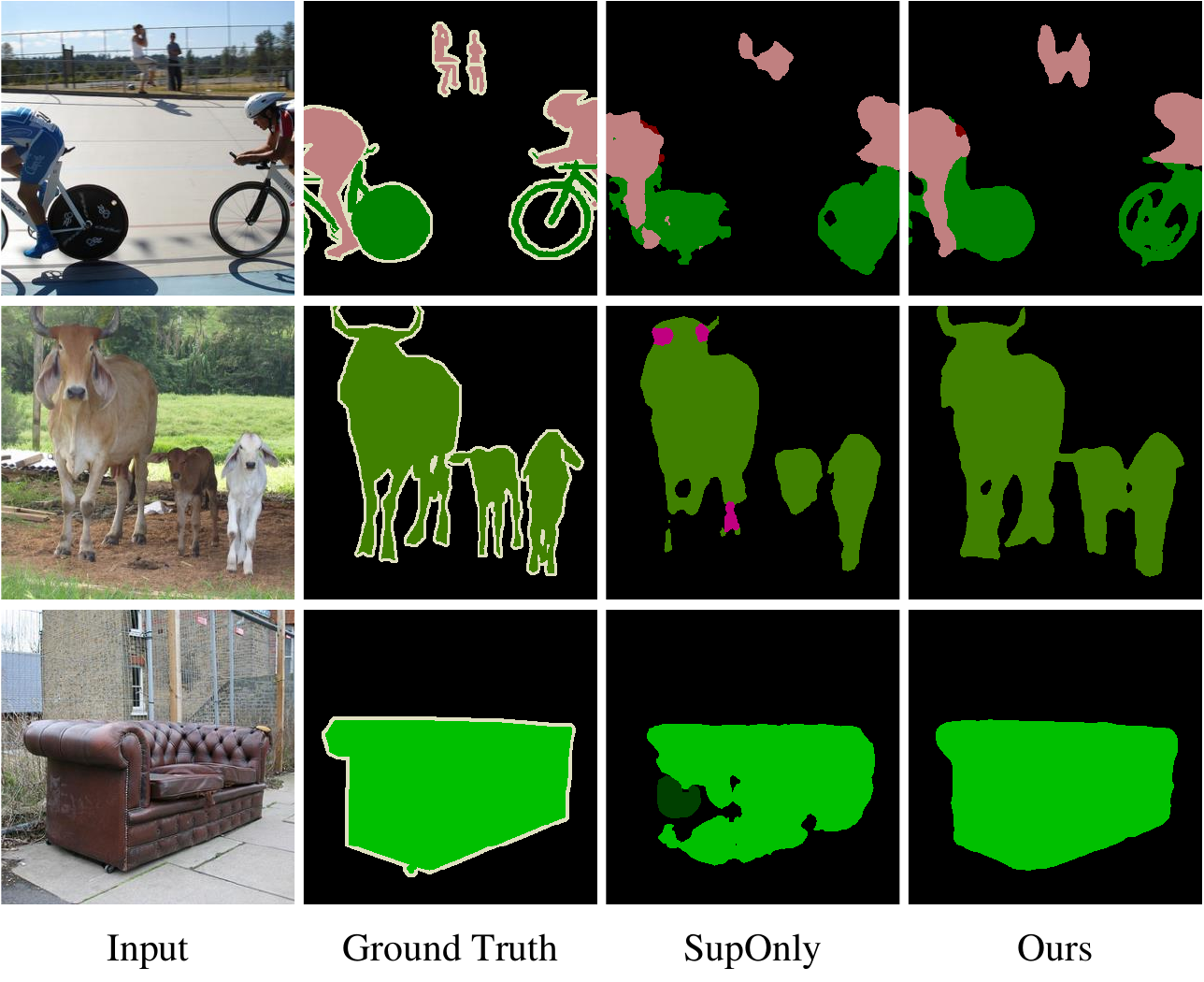}
	\caption{\textbf{Qualitative results} from Pascal VOC 2012.}
	\label{fig:8}
\end{figure}
\subsubsection{Qualitative Results.}\label{section:4.4}
Fig.~\ref{fig:8} visualizes some segmentation results on PASCAL VOC 2012. The supervised results display the bad accuracy caused by the limited labeled training samples. For example, in the 2-nd row, the supervised baseline mislabels the cow as the horse in many pixels. While our method successfully corrected the wrong annotation. Besides, the segmentation labeled by our method is more exquisite than the supervised-only method.

\section{Conclusions}

In this paper, we propose a new cross supervision based semi-supervised semantic segmentation approach, uncertainty-guided self cross supervision. Our method achieves self cross supervision by imposing the consistency between the subnetworks of a multi-input multi-out model. In order to alleviate the problem of noise accumulation and propagation in the pseudo label, we proposed uncertainty-guided learning, utilizing the uncertainty as guided information to reduce the effects of wrong pseudo labeling. Experiments show our approach dramatically reduces training costs and achieves powerful competitive performance.

\clearpage

\bibliographystyle{unsrtnat}
\bibliography{references}  %%% Uncomment this line and comment out the ``thebibliography'' section below to use the external .bib file (using bibtex) .





% \end{thebibliography}

\end{document}